\documentclass[conference]{IEEEtran}
\IEEEoverridecommandlockouts
\usepackage{cite}
\usepackage{amsmath,amssymb,amsfonts}
\usepackage{algorithmic}
\usepackage{graphicx}
\usepackage{textcomp}
\usepackage{xcolor}
\usepackage{url}
\usepackage{caption,tabularx,booktabs}
\def\BibTeX{{\rm B\kern-.05em{\sc i\kern-.025em b}\kern-.08em
    T\kern-.1667em\lower.7ex\hbox{E}\kern-.125emX}}
\begin{document}
\newcolumntype{Y}{>{\centering\arraybackslash}X}
\title{Machine Learning Based IoT Adaptive Architecture for Epilepsy Seizure Detection: Anatomy and Analysis\\
}

\author{\IEEEauthorblockN{Zag ElSayed, Murat Ozer, Nelly Elsayed}
\IEEEauthorblockA{\textit{School of Information Technology} \\
\textit{University of Cincinnati}\\
Ohio, United States \\
\{elsayezs, ozermm, elsayeny\}@ ucmail.uc.edu}
\and
\IEEEauthorblockN{Ahmed Abdelgawad}
\IEEEauthorblockA{\textit{School of Engineering and Technology} \\
\textit{Central Michigan University}\\
Michigan, United States \\
abdel1a@cmich.edu}
}

\maketitle

\begin{abstract}
A seizure tracking system is crucial for monitoring and evaluating epilepsy treatments. Caretaker seizure diaries are used in epilepsy care today, but clinical seizure monitoring may miss seizures. Monitoring devices that can be worn may be better tolerated and more suitable for long-term ambulatory use. Many techniques and methods are proposed for seizure detection; However, simplicity and affordability are key concepts for daily use while preserving the accuracy of the detection. In this study, we propose a versal, affordable noninvasive based on a simple real-time k-Nearest-Neighbors (kNN) machine learning that can be customized and adapted to individual users in less than four seconds of training time; the system was verified and validated using 500 subjects, with seizure detection data sampled at 178 Hz, the operated with a mean accuracy of (94.5\%).
\end{abstract}


\section{Introduction}
When brain cells breakdown and sends sabotaged electrical signals, it is called a seizure, in some cases different areas of human body are affected by such symptoms. Seizures are very common syndrome and affects more than 40\% of the population~\cite{ref1},~\cite{ref2}, unfortunately, for a variety of reasons, some people can experience them more frequently (diagnosed with Epilepsy). In many cases, seizures can be treated, especially if the underlying cause is known.
Classically, epilepsy is considered to be a syndrome of repeated seizure, however there is more of depth in the difference, as seizure falls in two categories. First, Provoked Seizure, they take place as a result of additional situations or conditions (high fevers, alcohol or drug withdrawal, low blood sugar), responsible for 25\% to 30\% of all seizures. Second, Unprovoked Seizures, they happen when a person's brain is more likely to develop spontaneous seizures and are not indicators of any present medical condition or environment. Additionally, epilepsy is diagnosed when a patient has at least two unprovoked seizures. Moreover, one unprovoked seizure increases the likelihood of at least one more in the subsequent ten years. Epilepsy cannot be determined by a doctor based only on provoked seizures~\cite{ref2}.
Seizure can cause problems with sleeping, thinking, and memory, or socializing with others. Seizure is responsible for (55\%) of sudden unexpected death in epilepsy (SUDEP) annually, as well as trauma, drowning, or other known causes~\cite{ref3}. Sudden unexpected death in epilepsy (SUDEP) is a rare but serious problem, the stages responsible for SUDEP are illustrated in Fig.~\ref{Figure1}. It is most common in people between the ages of 20 and 45, and is more common in males and those with childhood-onset epilepsy. It is generally thought that a seizure can cause changes in the brain and body that can lead to respiratory and cardiac problems, which may ultimately result in death. There is a higher odds ratio for SUDEP in these groups, and it is also more common in people who have epilepsy surgery~\cite{ref4}. Studies from the United States and Europe have shown that there is a higher rate of SUDEP in populations with socioeconomic barriers to care, such as lack of employment, lack of access to medications and other treatments, and increased distance from appropriate healthcare providers. However, based on recent population-based studies, there is no increased risk of SUDEP in patients with only absence or myoclonic seizures.
\begin{figure}[t]
		\centering
	\includegraphics[width=8cm,height=5cm]{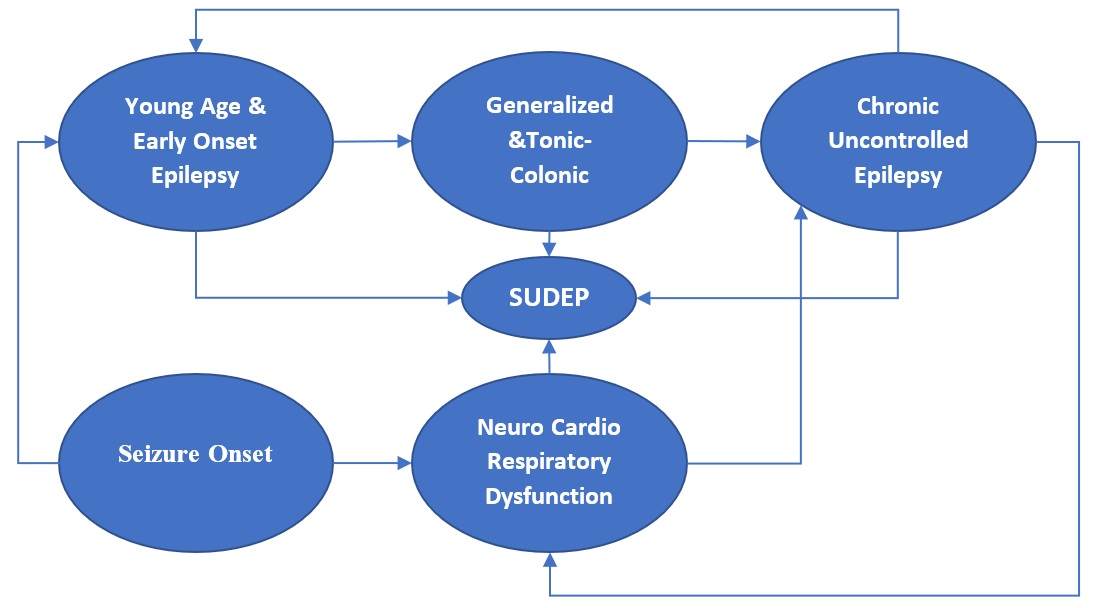}
	\caption{SUDEP Conditions and Triggers.}
	\label{figure1}
\end{figure}
The posterior thalamus plays a significant role in oxygen sensing and essential for breathing. A disruption of the thalamic-brain stem link in high-risk due to its involvement with respiratory failure in SUDEP and leads to asystole~\cite{ref5, ref6} is shown in Fig.~\ref{figure2}

\begin{figure}[t]
		\centering
	\includegraphics[width=6cm,height=6cm]{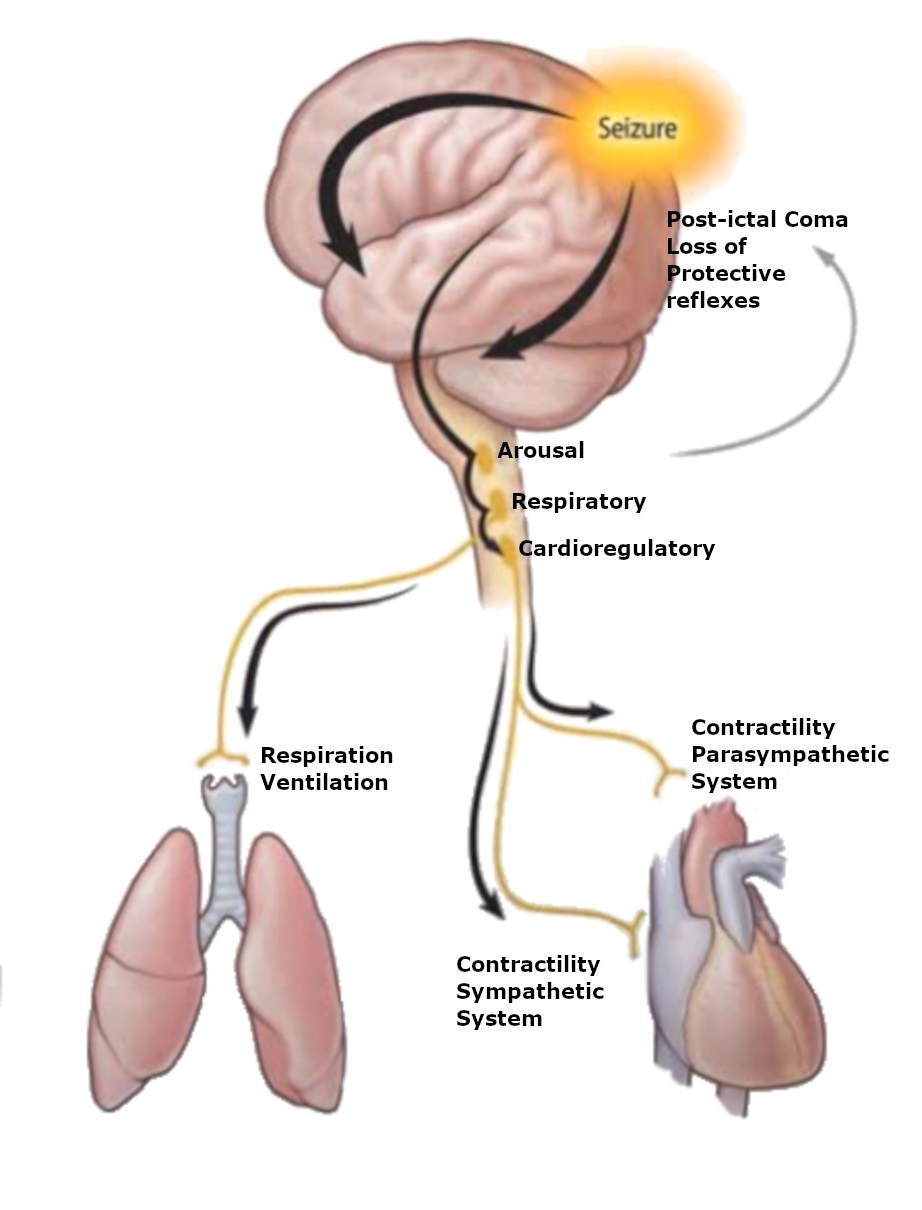}
	\caption{The effect of Seizure on Apnea, Arrythmia and Asystole.}
	\label{figure2}
\end{figure}

SUDEP it is a serious and potentially life-threatening complication of the condition, and people with epilepsy should be aware of the risks and take steps to minimize them, such as avoiding triggers for seizures, and seeking medical attention promptly if they experience any change. Thus, early detection of seizure via a wearable, affordable, and daily life usable technology can be a lifesaving.

The k-Nearest-Neighbors (kNN), KNN classifies or predicts new data points based on their similar instance proximity to the training samples~\cite{ref7}. The main positive characteristics of kNN algorithm over parametric techniques (e.g., linear regression) that it does not have any assumption by data before implementing the case module, and the constant evolution the kNN via its instance base learning provides a high advantage for compromising the learning model structure to fit individual patient’s behavior easily and in a timely efficient manner, especially within the for unpredicted nature of epilepsy seizure. Thus, and non-invasive, power efficient, real-time, fast adaptive wearable hardware implementation was proposed in this paper for seizure prediction via efficient kNN for a wide range of epilepsy diagnosed patients. the proposed architecture was trained and tested via real data form the University Hospital Bonn from 500 individuals including artifacts mainly from cornea-retinal dipole, retinal dipole and eyelid movement. The proposed system was implemented on a porotype Spartan 6 FPGA, and on a miniature embedded system System-on-a-Chip (SoC) ESP8266 32 bits 80MHz, during the validation and verification steps the system performers at 95\% accuracy rate of detection with a selected value of $k$. 

\section{Background}

Seizures result from abnormal electrical activity in the brain, often characterized by synchronous and excessive firing of a group of neurons~\cite{ref8}. This abnormal activity disrupts the normal functioning of the brain, leading to various clinical manifestations.
Neurons communicate with each other through electrical signals, which are regulated to maintain a balance between excitation and inhibition. In seizures, there is an imbalance favoring excessive excitation or reduced inhibition, resulting in a state of hyperexcitability~\cite{ref9}. Hyperexcitability can arise from various factors, including genetic predisposition, brain injury, infections, or metabolic imbalances.

\subsection{Brain Regions Involved}
Seizures can originate from specific brain regions or involve widespread neuronal networks~\cite{ref10}. The brain regions commonly associated with seizures include: Focal Seizures and Generalized Seizures.

Focal (Partial) Seizures can be further classified based on the affected region:
\begin{itemize}
	\item \textit{Focal (Partial) Temporal Lobe Seizures: }Originating from the temporal lobe cause complex behaviors, altered consciousness, and memory disturbances.
	\item \textit{Focal Frontal Lobe Seizures:} Arising from the frontal lobe, manifest as motor symptoms, such as repetitive movements or behaviors.
	\item \textit{Parietal Lobe Seizures:} Involving the parietal lobe, causes sensory disturbances, such as tingling, numbness, or hallucinations.
	\item \textit{Focal Occipital Lobe Seizures:} Originating from the occipital lobe, primarily affect vision, leading to visual hallucinations or blindness.	
	\item \textit{Generalized Seizures:} involve both cerebral hemispheres and typically have no focal onset. They include various subtypes as shown in Table.~\ref{table1}
\end{itemize}	

\begin{table}[htbp]
	\caption{Generalized Seizures Subtypes}
	\centering
		\begin{tabular}{|l|l|}
			\hline
			\textbf{Seizure}&\textbf{Description} \\
      		\hline
			Absence &Brief lapses in consciousness and staring spells, often\\
			& seen in children.\\
			\hline
			Tonic-Clonic &	Formerly known as \textit{grand mal} seizures, involve loss of \\
			&consciousness, stiffening (tonic), followed by \\
			&jerking movements (clonic).\\
			\hline
			Myoclonic&	Brief, sudden, and involuntary muscle \\
			&jerks that affect parts or the entire body\\
			\hline
			Atonic &	A drop attacks, cause a sudden loss of muscle tone,\\
			& leading to falls or head drops\\
			\hline
					\end{tabular}
		\label{table1}
\end{table}

\subsection{Neural Networks and Spread of Seizures}

Seizures can spread from the focal onset region to involve other brain regions, leading to a more generalized seizure, Fig.~\cite{figure3}. This propagation can occur through interconnected neural pathways and synchronized neural activity. The spread of seizures can result in the involvement of diverse functions and cause various symptoms depending on the affected brain regions~\cite{ref11}.

\begin{figure}[t]
		\centering
	\includegraphics[width=6cm,height=6cm]{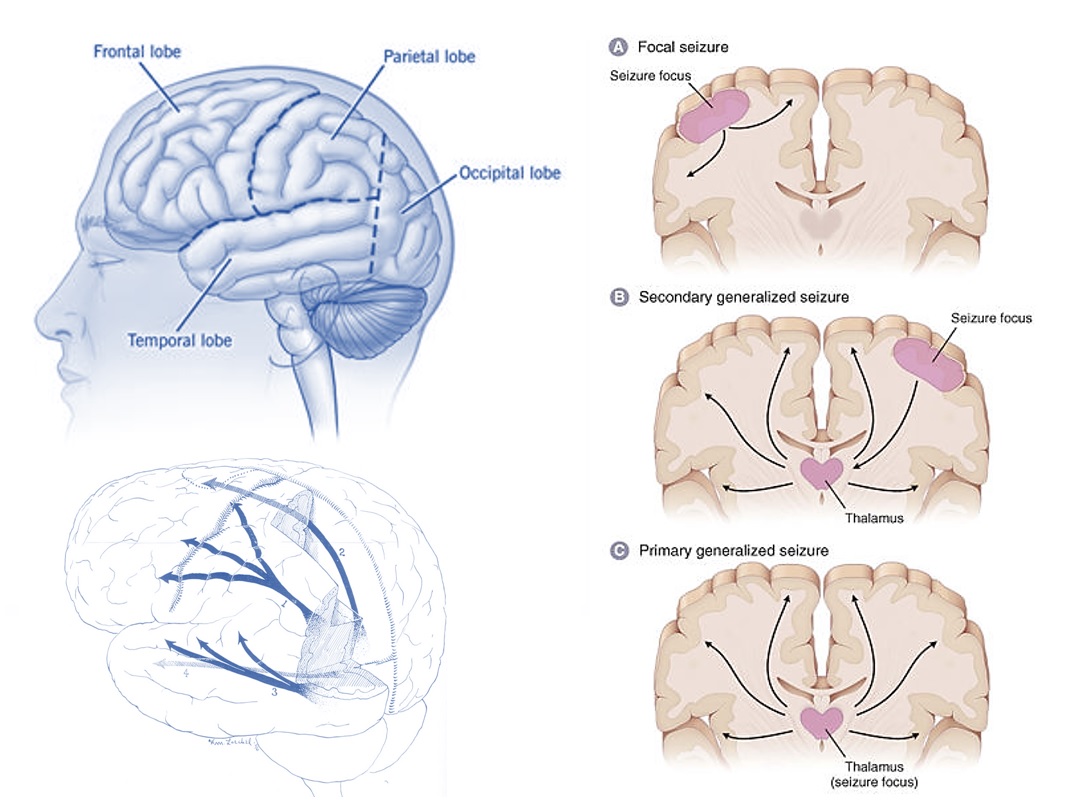}
	\caption{The Spread od Seizure im Focal, Secondery and Primary Generalized Seizures.}
	\label{figure3}
\end{figure}

Seizure propagation often follows specific patterns, depending on the brain networks involved. These networks can be anatomically connected regions or functional networks that share similar functions. Such as: Structural Connectivity and or Functional Connectivity~\cite{ref12}.  Seizure propagation can occur through several mechanism such as:

\begin{itemize}
	\item \textit{Synaptic Transmission:} during seizures, the abnormal electrical activity can enhance synaptic transmission, allowing the seizure activity to spread to adjacent neurons.
	\item \textit{Axonal Pathways: }The brain consists of intricate networks of axons. seizure activity can propagate along these axonal pathways, spreading to anatomically connected regions.
	\item \textit{Gap Junctions:} allow direct electrical coupling between adjacent neurons in the hippocampus and the gap junctions enabling the spread of electrical activity between neurons, promoting the synchronization of neuronal firing.
	\item \textit{Neurotransmitters:} Abnormal levels of neuro transmitters, such as glutamate, promote the spread of seizures by increasing neuronal excitability and synchrony.
\end{itemize}
Analyzing the mechanisms and brain regions involved in seizures is crucial for accurate detection of epilepsy. Electroencephalography (EEG) can be used capture and analyze the abnormal electrical activity associated with seizures, helping in diagnosis and treatment planning.

\subsection{Electroencephalography for Seizure Detection}
EEG is a non-invasive technique used to record and measure the electrical activity generated by the brain. In an EEG recording, electrodes are placed on the scalp at specific locations according to an internationally standardized (e.g., 10-20 system). These electrodes are typically made of metal or conductive gel and are connected to an EEG amplifier~\cite{ref13}. Additionally, EEG signals may contain unwanted artifacts, such as muscle activity (electromyographic activity), eye movements. electrode artifacts. skin/sweat artifact or electrical noise from the environment. To improve the quality of the EEG signal, filters are usually applied to eliminate these unwanted frequencies, solutions are discussed in~\cite{ref14}.

\begin{figure}[t]
		\centering
	\includegraphics[width=8cm,height=5cm]{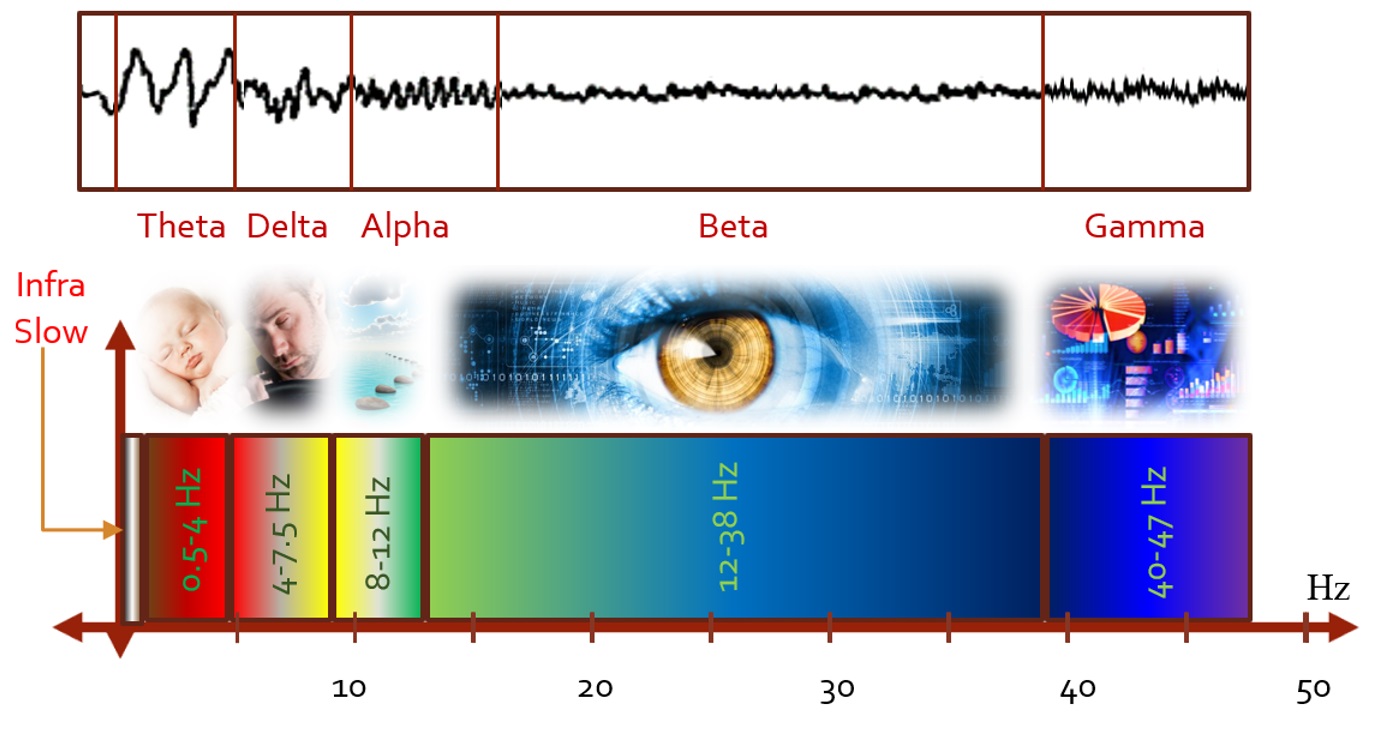}
	\caption{EEG Frequency Bands with specific brain states and activities.}
	\label{figure4}
\end{figure}

The recorded EEG signal consists of different types of waves that represent different patterns of brain activity. These waves are categorized into frequency bands, each associated with specific brain states and activities, shown in Fig.~\ref{figure4}.

Delta Waves (0.5-4 Hz) typically observed during deep sleep or in pathological conditions. Theta Waves (4-8 Hz): occur during drowsiness, meditation, or certain types of brain disorders. Alpha Waves (8-13 Hz) present during relaxed wakefulness, or closed eyes. Beta Waves (13-30 Hz) associated with active thinking, concentration, and alertness. Gamma Waves ($\mathrm{>}$30 Hz) observed during cognitive processing.

\subsection{EEG Seizure Patterns}
Interictal epileptiform discharges (IED) are abnormal electrical patterns observed between seizures~\cite{ref15}. These discharges can serve as markers of epileptic activity and provide valuable diagnostic information. Some common types of IEDs include: \textit{Spike and Sharp Waves} with high-amplitude deflections typically have a duration of less than 200 milliseconds and are characterized by a rapid onset and offset with a pointed morphology, while sharp waves have a broader appearance. \textit{Sharp and Slow Wave Complexes} consist of a sharp wave followed by a slow wave component. The sharp wave component is similar to a spike or sharp wave, while the slow wave component has a lower frequency and longer duration, this often indicates a more prolonged abnormal electrical activity in the brain. \textit{Poly-spikes} are characterized by the presence of multiple spike-like waveforms occurring in rapid succession in specific epilepsy syndromes, such as juvenile myoclonic epilepsy. Sharp and Slow Wave Bursts are clusters of sharp waves followed by slow waves that can last for several seconds and often occur repeatedly within a short period.

On the other hand, EEG Seizure pattern can be found in Ictal events, they represent the actual seizure activity in real-time and provide insights into the dynamic changes occurring in the brain during a seizure. Some notable ictal EEG patterns include: \textit{Low Voltage Fast Activity:} a sudden increase in the frequency of the EEG activity, often accompanied by a decrease in amplitude ranges from 10 to 30 Hz and may be seen in the initial phases of seizures. \textit{High Amplitude Slow Waves }often have a frequency of less than 4 Hz and reflect synchronous neuronal depolarization and hyperexcitability. \textit{Spike and Spike-and-Wave Complexes }by the presence of repetitive spikes, sharp waves, or spike-and-wave complexes indicative of intense and synchronized neuronal activity.

EEG is a convenient and relatively inexpensive way to demonstrate the physiological manifestations of epilepsy. However, the EEG has a number of limitations. Electrical activity recorded by electrodes placed on the scalp or surface of the brain mostly reflects summation of excitatory and inhibitory postsynaptic potentials in apical dendrites of pyramidal neurons in the more superficial layers of the cortex. Additionally, limited seizure patterns resolution as well as the fact that EEG primarily measures electrical activity from the cortical surface, limiting its spatial resolution. Deep-seated brain activity (e.g., mesial temporal structures) may not be fully captured by scalp EEG challenging localizing the onset zones. Moreover, EEG has a limited continuous monitoring capability.

Despite these challenges and limitations, EEG remains a valuable tool in seizure detection, diagnosis, and monitoring for many individuals with epilepsy. Especially, when paired with advanced multimodal DSP sophisticated machine learning techniques to enhance usability, accuracy and reliability of EEG-based seizure detection systems. Thus, we propose a novel, affordable simple real-time kNN machine learning architecture that adapts to individual users in less than four seconds, and detects seizure at (94.5\%) accuracy.

\section{Methodology and Results}
\begin{figure}[t]
		\centering
	\includegraphics[width=8cm,height=7cm]{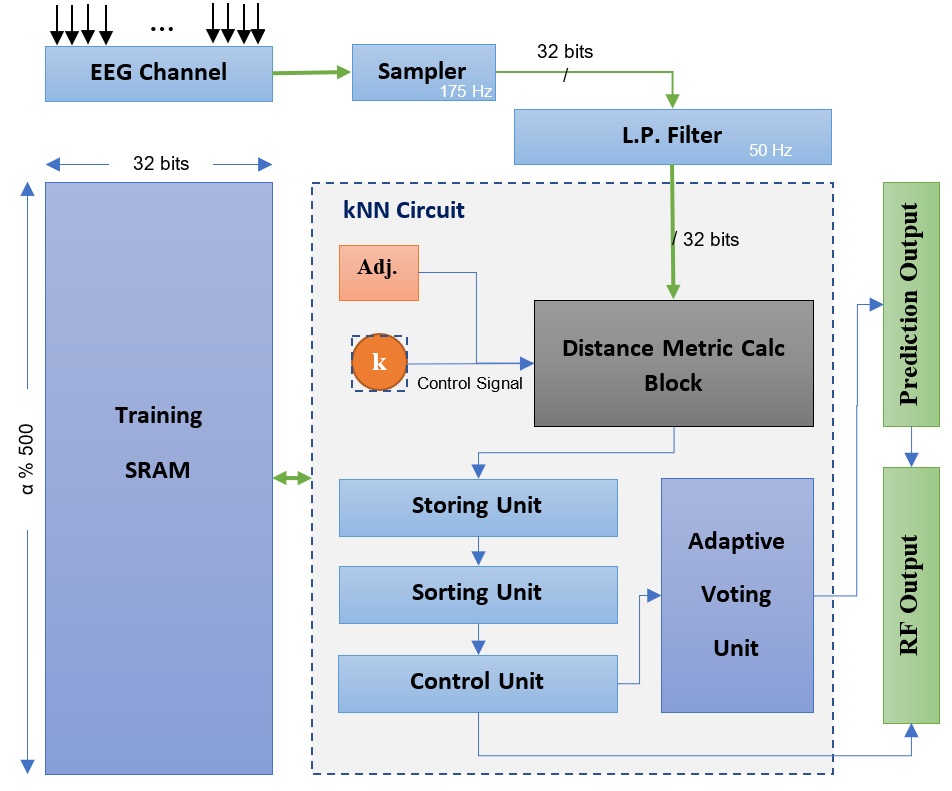}
	\caption{The proposed System architechure bolck diagram.}
	\label{figure5}
\end{figure}

There are many state-of-the-art technologies proposed for seizure detection via EEG, such as, and not limited to, convolutional neural networks (CNNs) and recurrent neural networks (RNN)~\cite{ref16}, Long Short-Term Memory (LSTM) Networks~\cite{ref17}, and responsive neurostimulation (RNS) transcranial magnetic stimulation (TMS)~\cite{ref18}. However, their technologies require enormous computing power, rich, long training phase data sets, and specialized data acquisition equipment, respectively. Thus, in this study, we propose a hardware implementation of a simplified kNN-based seizure detection architecture featured by simple and short-term reequipment for the individual patient signals learning phase, higher than average accuracy, and an easy wearable system for everyday use. The proposed system design shown in Fig.~\ref{figure5}, due to the size limitation we the unit component are describe briefly.

The system has four input EEG dry electrodes placed on the patient's skull (based on the type of the diagnosed seizure type/zones); the input analog input is sampled at 174Hz, then passed to the Low Pass filter before being processed by the kNN Circuit Unit (kC) that was inspired by the fast kNN implementation in~\cite{ref19} and reduced to fit EEG data sophisticated nature. The kC consists of five (5) stages. Distance Metric Calc Block (DMCD) computes the distances between the test and training data points using Euclidean distance and a performance metric.
\begin{equation}
	d(x_i, x_j) = \sqrt{\sum_{r=1}^{n} (a_r(x_i)-a_r(x_j))^2}
\end{equation}

\begin{figure}[t]
	\centering
   \includegraphics[width=8cm,height=5cm]{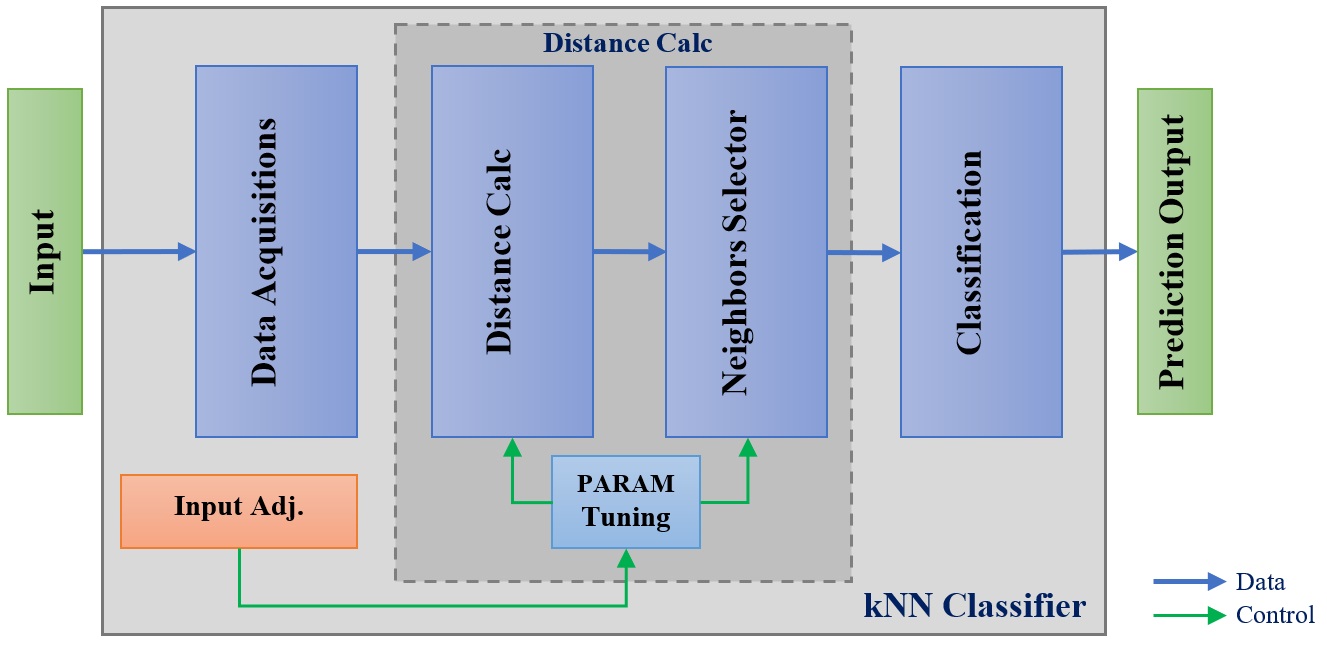}
	\caption{The proposed adjusted kNN algorithm logical block diagram.}
	\label{figure6}
\end{figure}

Where $d(x_i, x_j)$ is Euclidean Distance, $(x_i)$ is ith record, $(x_j)$ is jth record, $(a_r)$ is $\mathrm{r^{th}}$ record in the data sample space of size \textit{n, i,j is 1 , 2, 3,..., n}.  The $k$ in kNN determines the number of points used as a reference to classify the input activity, shown in Fig.~\ref{figure6}. The Storage Unit (SU) is activated during the operation's training and testing phases. It is responsible for storing and feedback on the training SRAM based on the preset parameters of $k$ and the input prefaces training sample size $(\alpha)$, which determines the memory usage and the area needed for the Training Data storage.

Although kNN technique is quite simple in performance and does not require a lot of computing power, it requires expensive comparison calculations to compare (unlabeled) data points to other labeled data points (i.e., training data); hence it might seem to be not ideal to be used on stand-alone wearable devices for seizure detection. However, in our proposed method, we use a state-of the-art arithmetic logic unit (ALU) that enables the usage of the 1's complement digital presentation to chive $O(1)$ comparison unit~\cite{ref20}.  

Then the Control Unit (CU) selects the k nearest neighbors based on the computed distance using \textit{O(k)} hardware sorting algorithm circuits, then passes the calculations in case of detection to directly the Output Unit or to the Adaptive Voting Unit (AVU) where a majority voting scheme to determine the class label of the test data point while keeping track of the class labels of the k nearest neighbors and select the label with the highest count. Finally, The Output Units provide the classification result for the detection as well as transmit the results to the connected external device via nRF52833 WLCSP with 2.402 MHz - 2.480 MHz (i.e., Low Power 802.15 range) with a transmission power of + 8 dBm. The hardware specifications are shown in Table.~\ref{table2}.

\begin{table}[htbp]
	\caption{Microprocessor and RF Board Hardware Specs.}
	\begin{center}
		\begin{tabular}{|l|l|}
			\hline
			\textbf{Category}&\textbf{Parameters} \\
	\hline
			
		Protocols&	802.11 802.15 1\\
		\hline
		CPU&Tensilica L106 32-bit processor\\
		\hline
		Memory&	32 KiB instruction, 80 KiB user data\\
		\hline
		V, A&	2.5 V \~ 3.6 V, 80 mA\\
		\hline
		Temp Range&	40$^\circ$ to +105$^\circ$ C\\
		
			\hline
		
		\end{tabular}
		\label{table2}
	\end{center}
\end{table}

\begin{figure}[t]
	\centering
\includegraphics[width=8cm,height=11cm]{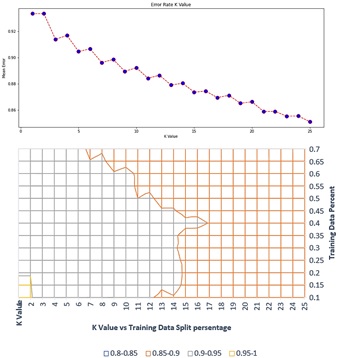}
	\caption{The accurace of the system at k=3 and the selection value for ($\alpha$).}
	\label{figure7}
\end{figure}

During the verification and validation steps, different values of k were tested against the sample size the sample size $(\alpha)$ and the results are shown in Fig.~\ref{figure7}. Where the best value of k was selected (k=3) and ($\alpha$ = 30).

The result of this experiment is compared to the closet state-of-the-art regarding the algorithm mused, the implementation, the dataset involved and the result accuracy, is shown in Table~\ref{table3}. 

\begin{table}[htbp]
	\caption{Microprocessor and RF Board Hardware Specs.}
	\begin{center}
		\begin{tabular}{|l|c|c|c|c|c|}
			\hline
			\textbf{Architecture}&\textbf{Algorithm}&\textbf{Dataset}&\textbf{SW}&\textbf{HW}& \textbf{Accuracy}\\
		\hline 
			Mursalin et al.&	Genetic&	CHB-MIT	& $\surd$	& &	97.14\%\\
			Hasan. et al.&	kNN&	Bonn DB&	$\surd$ 	& &	96\%\\
			Shoaib et al&	SVM&	MIT EEF&$\surd$ &		&93\%\\
			Kaveth et al&	SVM/RF&	CHB-MIT	&&	$\surd$ 	&92\%\\
			Choubey &	ANN+kNN&	Bonn DB	&&	$\surd$ 	&94\%\\
			Wei et al.&	CNN&	CHB-MIT	&&	$\surd$ 	&90.57\%\\
			\textbf{Proposed}&	kNN	&Bonn DB	&&	$\surd$ &	\textbf{94.5\%}\\

			\hline

		\end{tabular}
		\label{table3}
	\end{center}
\end{table}
\section{Conclusion}
Noninvasive seizure detection using EEG signals has witnessed significant advancements in understanding the underlying neural mechanisms and EEG patterns. Despite the EEG patterns and signal challenges, the proposed system provides a simple, adapt wearable architecture implementation that can play a significant role in achieving real-time and efficient seizure detection, reducing SUDEP, and holding promise for improving the lives of individuals living with epilepsy and advancing the field of seizure detection and management.

\bibliographystyle{ieeetr}
\bibliography{references_list}

\end{document}